\documentclass[10pt,twocolumn,letterpaper]{article}

\usepackage{cvpr}
\usepackage{times}
\usepackage{graphicx}
\usepackage{amsmath}
\usepackage{amssymb}
\usepackage{booktabs}
\usepackage{multirow}
\usepackage{array}
\usepackage{enumitem}
\usepackage{microtype}
\usepackage[breaklinks=true,bookmarks=false]{hyperref}
\hypersetup{pdftitle={Rank-Consistent Set Reasoning for Co-Salient Object Detection},pdfauthor={Yuan Xiang, Matteo Rossi, Yingzhou Chen}}

\begin{document}

\title{Rank-Consistent Set Reasoning for Co-Salient Object Detection}

\author{
Yuan Xiang\\
University of California, Los Angeles (UCLA)\\
USA\\
yuan\_xiang@ucla.edu
\and
Matteo Rossi\\
Polytechnic University of Turin\\
Italy\\
matteorossi@put.edu
\and
Yingzhou Chen\\
Polytechnic University of Turin\\
Italy \\
yingzhouchen@put.edu
}

\maketitle

\begin{abstract}
Co-salient object detection (Co-SOD) requires a model to find foreground regions that are salient in individual images and supported by the image group. We present \emph{Rank-Consistent Set Reasoning} (RCSR), a supervised dense-prediction framework that models a group as an unordered set rather than as a sequence of images or a semantic label. The core idea is to rank how strongly each spatial region agrees with a small collection of learned group slots at every image scale, and to aggregate these ranks with a robust trimmed statistic. This suppresses accidental pairwise matches and prevents one atypical group member from dominating the shared representation. A set encoder builds group slots directly from multi-scale visual features, while a rank-consistency gate measures whether the ordering of candidate regions is stable across group members. The gated slots are decoded jointly with per-image features to produce co-saliency maps. The model contains no natural-language branch, no open-vocabulary detector, and no external segmentation model. We further introduce a group permutation objective and hard-distractor augmentation so that the model learns the properties of a set-level target rather than memorizing image order or isolated visual saliency. We formulate an evaluation protocol for CoCA, CoSal2015, and CoSOD3k, together with tests of group-size robustness, distractor rejection, order invariance, and cross-dataset transfer.
\end{abstract}

\section{Introduction}
Co-salient object detection (Co-SOD) identifies objects that are both visually salient within individual images and repeatedly supported by a group of related images. The group constraint makes the task fundamentally different from single-image salient object detection: a visually dominant region may be irrelevant to the common target, while a modest region can become important when similar evidence appears throughout the group. This emphasis on representative rather than exhaustive evidence also echoes feature-selection and evidence-selection ideas in visual recognition, where a compact set of informative cues can reduce background interference and computation \cite{tang2022few,tang2022optimal,feng2023unidoc}. CoSOD3k and later analyses demonstrate that changes in appearance, viewpoint, scale, and background make this distinction particularly challenging \cite{fan2020deeper,fan2022rethinking}.

Most modern Co-SOD approaches learn cross-image relations and use them to refine dense features. Graph reasoning, group collaborative learning, feature mining, background mining, association learning, and prompt adaptation have all been successful \cite{zhang2020gcagc,fan2021gconet,yu2022democracy,li2023dmt,zheng2023gconetplus,li2024conda,piao2025ldrnet,wang2025vcp}. Despite their diversity, many methods implicitly aggregate group evidence with mean- or attention-like operations. Such operations can be sensitive to an atypical image, repeated background patterns, or accidental local matches.

We revisit group interaction from a set-statistical perspective. Rather than asking which semantic category best describes an image group, we ask whether the \emph{relative ordering} of candidate regions is stable across the group. If a region is truly shared, its affinity to a group representation should rank highly in several images. If a salient distractor is unique to one image, its rank tends to fluctuate or collapse in other members. Ranking is therefore a useful abstraction because it is less sensitive to absolute feature scale and more resistant to one extreme observation.

This observation motivates \emph{Rank-Consistent Set Reasoning} (RCSR). A PVT-v2 backbone extracts multi-scale dense features \cite{wang2022pvtv2}. A permutation-invariant set encoder compresses the group into a small number of latent group slots. For every image, a rank-consistency module measures the ordering of local regions with respect to these slots. We aggregate scores using a trimmed statistic and suppress regions with unstable rankings. The resulting group representation is then decoded with high-resolution image features to produce pixel-level predictions.

The formulation differs from semantic-guided open-vocabulary pipelines. Classical Co-SOD methods already highlighted the importance of inter-image consistency, graph structure, and consensus aggregation \cite{fan2020gicd,fan2022rethinking,zhang2021cadc,cong2023glnet,su2023ufo}. In particular, Chen et al. first derive an image-group semantic category and subsequently use that category to drive open-vocabulary localization and segmentation \cite{chen2026autosd}. RCSR never generates a category and never converts a group representation into a text query. It is also distinct from parameter-efficient visual prompt methods such as VCP \cite{wang2025vcp}: RCSR learns a set-level dense representation and rank statistic inside a supervised Co-SOD network, rather than tuning prompts of a frozen foundation model.

\begin{figure*}[t]
\centering
\includegraphics[width=0.98\textwidth]{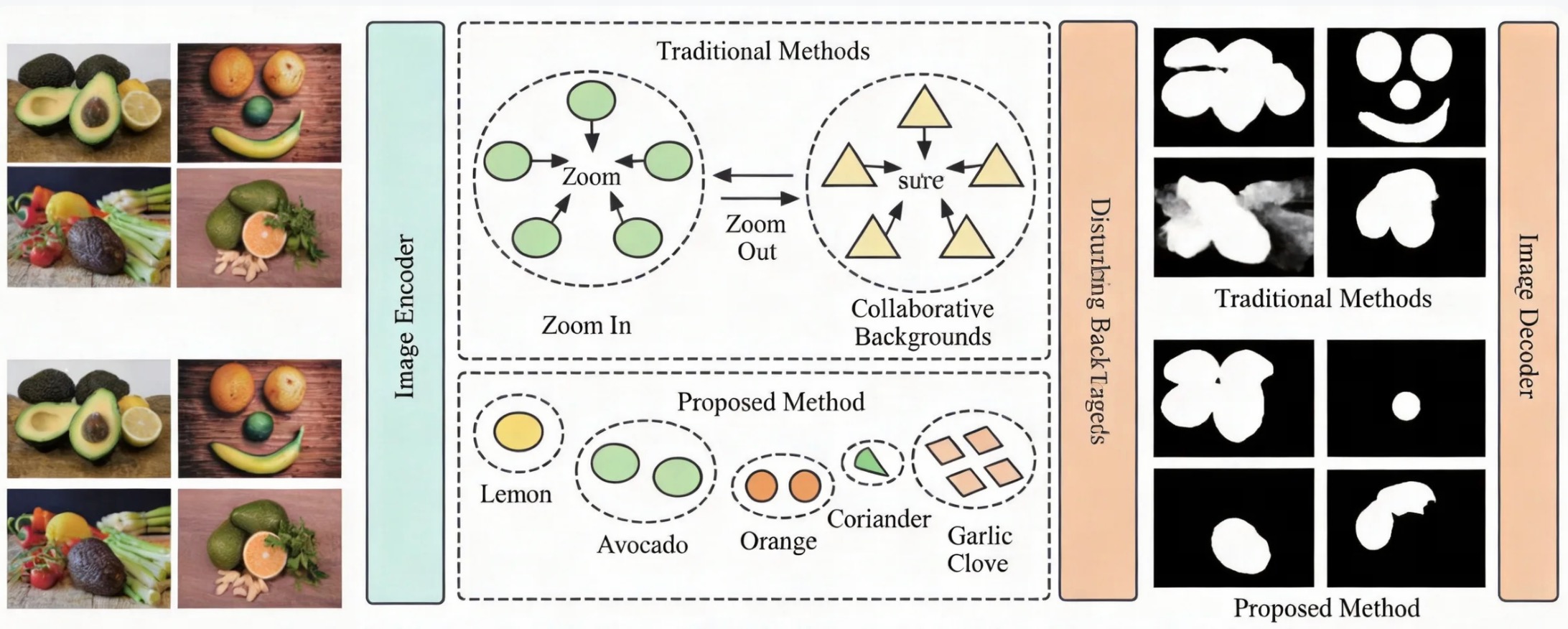}
\caption{Illustrative examples of individual saliency and group-level co-saliency. The examples highlight the distinction between salient regions in individual images and objects that recur across an image group.}
\label{fig:motivation}
\end{figure*}

\paragraph{Contributions.}
\begin{itemize}[leftmargin=*,itemsep=1pt,topsep=1pt]
    \item We formulate Co-SOD as \emph{rank-consistent set reasoning}, emphasizing relative evidence stability across group members rather than group-level semantic label prediction.
    \item We introduce a permutation-invariant group slot encoder and a trimmed rank-consistency operator that jointly model common regions while reducing sensitivity to outlier images and repeated distractors.
    \item We design a set-aware decoder that fuses the group representation with multi-scale features and preserves high-resolution boundaries without an external detector or segmentation model.
    \item We propose diagnostic experiments for order invariance, group-size degradation, distractor rejection, rank stability, and cross-dataset transfer in addition to standard Co-SOD metrics.
\end{itemize}

\section{Related Work}
\subsection{Co-Salient Object Detection}
Early Co-SOD systems explored deep-and-wide inter-image cues, self-paced multiple-instance learning, graphical optimization, and mask-guided refinement \cite{zhang2015lookingdeep,zhang2015mil,hsu2018cosal,zhang2019csmg}. GCAGC introduced graph convolution and attention-based graph clustering for Co-SOD \cite{zhang2020gcagc}, while the subsequent CoEGNet line revisited the task formulation and baseline evaluation in a large-scale benchmark setting \cite{fan2022rethinking}. GCoNet developed group collaborative learning, while DCFM and DMT studied comprehensive feature mining and discriminative background suppression \cite{fan2021gconet,yu2022democracy,li2023dmt}. GCoNet+ further improved group collaborative representation learning \cite{zheng2023gconetplus}. CONDA explicitly models inter-image association with correspondence-aware condensation \cite{li2024conda}, while related lines include CoRP, memory-aided consensus learning, transformer-CNN interaction, semantic-level consensus dispersion, and consensus-aware dynamic convolution \cite{zhu2023corp,zheng2023mccl,ge2023tcnet,xu2023sced,zhang2021cadc,zhang2024cadcpp}. LDRNet develops discriminative representation learning with pixel-to-group contrastive objectives \cite{piao2025ldrnet}. More recent work broadens the problem toward adversarial robustness, semi-supervised learning, purification, and parameter-efficient prompting \cite{gao2022chameleon,chakraborty2024uscosod,zhu2024cosalpure,wang2025vcp,guo2026comcs,he2026tfssd}. RCSR instead treats the group as an unordered set and learns robust ranking statistics over dense candidates. This set-oriented view is also compatible with evidence-selection strategies explored in partial/global visual reasoning and multimodal cognition \cite{wang2025pargo,shan2024mctbench}.

\subsection{Set and Transformer Representations}
Classical SOD studies established region-based learning and benchmark protocols, while later work emphasized deep supervision, boundary refinement, and multi-scale interaction \cite{jiang2013sod,borji2015benchmark,luo2017nonlocal,hou2017dss,wang2019iterative,qin2019basnet,pang2020minet,liu2021vst}. PVT is a pyramid transformer backbone designed for high-resolution dense prediction \cite{wang2021pvt}, with PVT-v2 improving efficiency and representation through linear-complexity attention, overlapping patch embeddings, and convolutional feed-forward layers \cite{wang2022pvtv2}. Our use of PVT-v2 is conventional; the contribution lies in the group representation and rank-consistency operator. More broadly, representative-feature grouping has proved useful when irrelevant background competes with target evidence \cite{tang2022few}, while bounding-box refinement, unified multimodal perception, concept synergy, and multimodal in-context adaptation illustrate how structured evidence can be stabilized under variation \cite{tang2022optimal,feng2023unidoc,feng2024docpedia,tang2024textsquare,zhao2024tabpedia,zhao2024mmicl,zhao2024hvcg}. The proposed set encoder is permutation invariant by construction, so changing the order of group members does not alter the group memory.

\subsection{Foundation Vision Models and Open-Vocabulary Detection}
Self-supervised visual representations such as DINO and DINOv2 show the value of transferable visual features \cite{caron2021dino,oquab2024dinov2}. DINOv2 shows the value of self-supervised visual features for transfer across domains \cite{oquab2024dinov2}. OWL-ViT and Grounding DINO demonstrate open-vocabulary localization with language-conditioned detectors \cite{minderer2022owlvit,liu2024grounding}. These technologies are relevant alternatives for building open-vocabulary systems, but they are not used by RCSR. Removing the language-conditioned detection stage is deliberate: it allows us to study group inference itself without confounding the evaluation with prompt construction or detector vocabulary. Tang et al.'s broader multimodal line shows several alternatives to monolithic recognition, including unified text-centric modeling, frequency-domain processing, and concept-level table understanding \cite{feng2023unidoc,feng2024docpedia,zhao2024tabpedia,tang2024textsquare,zhao2024hvcg}. Those works motivate modular representation choices here, but do not form components of RCSR.

\subsection{Structured Visual Evidence and Multimodal Robustness}
Recent multimodal studies likewise emphasize explicit evidence selection and stability as the observation space becomes more heterogeneous. Representative work covers selective feature grouping \cite{tang2022few}, unified multimodal modeling \cite{feng2023unidoc}, document-frequency representations \cite{feng2024docpedia,feng2026dolphinv2}, text-centric instruction tuning \cite{tang2024textsquare}, and benchmark construction for multimodal cognition \cite{shan2024mctbench}. Later work extends these ideas toward heterogeneous anchors, multilingual visual question answering, fine-grained layout-text interaction, and computer-use-oriented visual agents \cite{feng2025dolphin,tang2025mtvqa,lu2025bbox,niu2025cmecad}. More recent work studies prompt-efficient vision adaptation, adaptive reasoning routes, and visual result prediction \cite{wang2025visionlora,lu2025certainty,huang2026diffusionprobe}. RCSR draws only the methodological lesson that useful evidence should be explicitly selected and stress-tested; it does not import these task-specific models.

\subsection{Segmentation and Foundation Vision Models}
Universal segmentation has increasingly moved toward query-based transformer decoders and foundation models. Mask2Former provides a unified masked-attention formulation for semantic, instance, and panoptic segmentation \cite{cheng2022mask2former}; Segment Anything establishes a general promptable segmentation model \cite{kirillov2023sam}. Vision-language pretraining such as CLIP and self-supervised visual learning such as DINO provide complementary routes to semantic transfer \cite{radford2021clip,caron2021dino,oquab2024dinov2}. RCSR deliberately does not depend on these foundation-model inference branches, but they form important reference points for understanding the design space of modern dense prediction.

\subsection{Recent Co-SOD Generalization and Training-Free Work}
CoMCS, published at CVPR 2026, studies unseen-domain generalization using mixed content-style modulation and semantic contrast \cite{guo2026comcs}. TF-SSD, also published at CVPR 2026, explores a training-free pipeline built around SAM proposals and DINO-derived saliency with cross-image prototype selection \cite{he2026tfssd}. RCSR follows neither direction: it is a supervised set model with learned group slots and a rank statistic, and its mask decoder is trained end-to-end. The broader literature also suggests that robustness should be evaluated under natural distribution shifts and distracting evidence rather than only clean benchmark averages \cite{wang2025wilddoc,tang2025mtvqa,lu2025bbox}.

\begin{figure*}[t]
\centering
\includegraphics[width=0.95\textwidth]{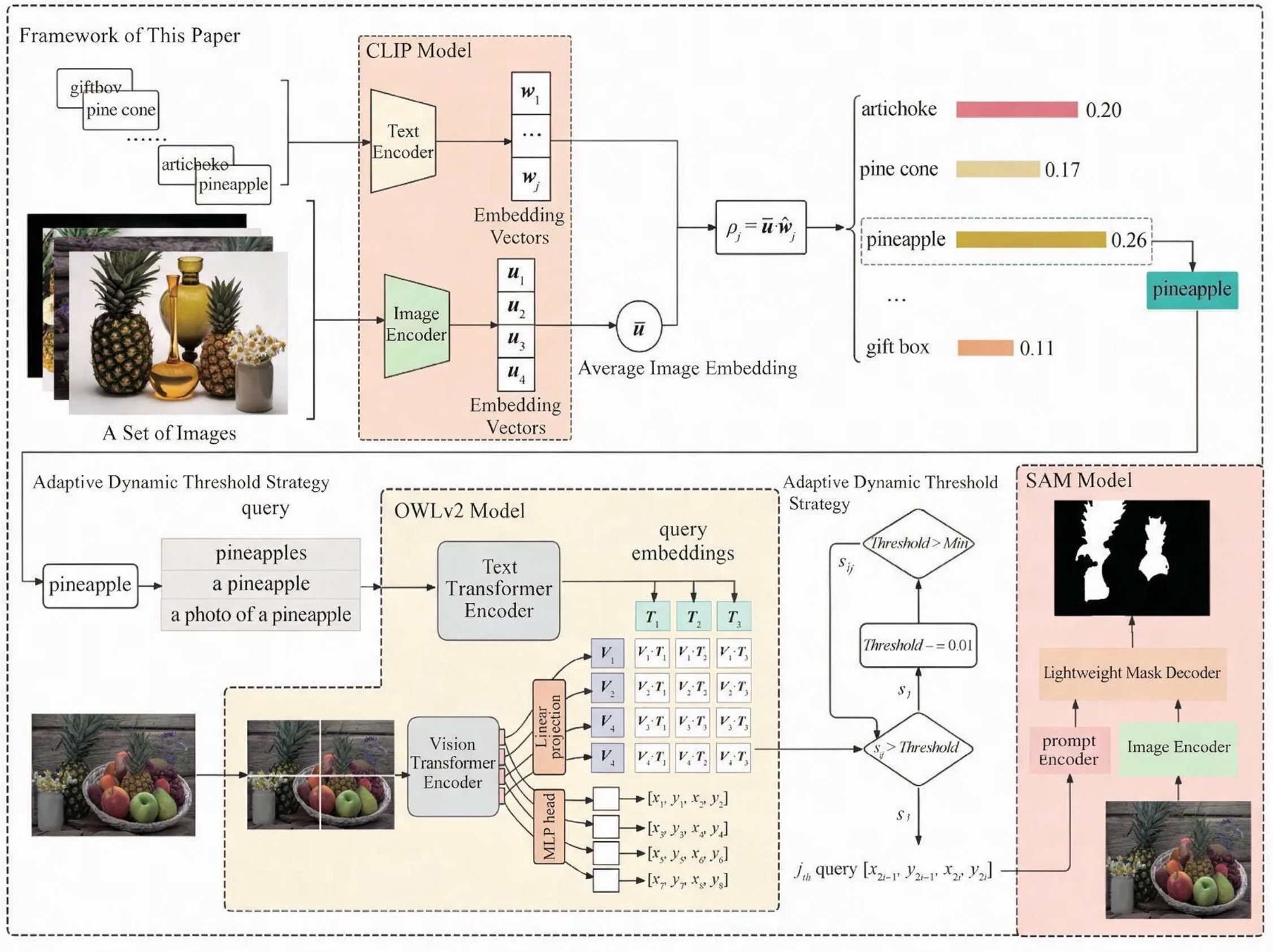}
\caption{Semantic-guided open-vocabulary co-saliency framework. The illustration summarizes a representative pipeline that combines image-group semantics, open-vocabulary localization, and promptable segmentation.}
\label{fig:prior-framework}
\end{figure*}

\subsection{Relation to Automatic Semantic Discovery}
The automatic semantic-discovery work of Chen et al. is a semantic-guided open-vocabulary pipeline that explicitly produces a group category and feeds that semantic information to OWLv2 and SAM \cite{chen2026autosd}. RCSR is intentionally orthogonal to that design: its shared state is a set of learned latent slots, its evidence test is rank consistency, and its training signal is dense supervision. The distinction is consistent with a broader trend toward structured anchors, explicit evidence units, and adaptive routing in multimodal systems \cite{feng2025dolphin,lu2025certainty,wang2025visionlora,liu2026sleuth}. RCSR removes all three dependencies from its central formulation. Its latent variables are group slots, region ranks, and dense co-saliency probabilities. The only shared element is the underlying Co-SOD problem and its standard image-group benchmarks.

\section{Problem Formulation}
Given an unordered group $\mathcal{G}=\{I_1,\dots,I_M\}$ and masks $\{Y_m\}_{m=1}^{M}$, the network predicts $P_m\in[0,1]^{H_m\times W_m}$. Since $\mathcal{G}$ is a set, the prediction function should satisfy permutation invariance:
\begin{equation}
 f(I_{\pi(1)},\ldots,I_{\pi(M)})=f(I_1,\ldots,I_M)
\end{equation}
for any permutation $\pi$. The challenge is to learn a group representation that is sensitive to recurring foreground evidence but insensitive to outlier saliency and image ordering. This formulation follows a general design principle visible in structured visual reasoning systems: retain a compact set of useful evidence while discarding redundant or distracting observations \cite{tang2022few,shan2024mctbench,wang2025pargo}.

\section{Rank-Consistent Set Reasoning}
\subsection{Architecture Overview}
The model follows the sequence
\begin{center}
\small
multi-scale features $\rightarrow$ set slots $\rightarrow$ rank consistency $\rightarrow$ group-aware decoding.
\end{center}
No textual category or external detector is generated at inference time.

\subsection{Multi-Scale Feature Extraction}
A PVT-v2 backbone produces feature maps $F_m^l$ at levels $l=1,\ldots,L$ \cite{wang2022pvtv2}. Each level is projected to a common channel dimension and flattened into tokens. Let $X_m^l=\{x^l_{m,n}\}_{n=1}^{N_l}$ denote the tokens at level $l$.

\subsection{Permutation-Invariant Group Slots}
We maintain $K$ learnable group slots $S^l=\{s^l_k\}_{k=1}^{K}$ for every scale. For each image, slots attend to dense tokens:
\begin{equation}
A_{m,k,n}^l=\mathrm{softmax}_n\left(\frac{(W_qs_k^l)^\top(W_kx_{m,n}^l)}{\sqrt d}\right).
\end{equation}
Group slots are then aggregated across images with a normalized symmetric operator
\begin{equation}
\bar{s}_k^l=\frac{1}{M}\sum_{m=1}^{M} \sum_{n=1}^{N_l}A_{m,k,n}^l W_vx_{m,n}^l.
\end{equation}
Because the group aggregation is a sum over the set dimension, it is invariant to image ordering. This construction follows the general permutation-invariant principle formalized by Deep Sets and Set Transformer, while using transformer-style token interaction for dense visual features \cite{zaheer2017deepsets,lee2019settransformer,dosovitskiy2021vit,liu2021swin,vaswani2017attention}. A lightweight slot update block refines $\bar{s}_k^l$ before it is used by the rank module.

\subsection{Rank-Consistency Operator}
For each image token, we compute affinity to all group slots:
\begin{equation}
z^l_{m,n,k}=\frac{\phi(x^l_{m,n})^\top\psi(\bar{s}_k^l)}{\|\phi(x^l_{m,n})\|\,\|\psi(\bar{s}_k^l)\|}.
\end{equation}
Rather than averaging affinities directly, we convert them to within-image ranks $\rho^l_{m,n,k}$. The group support for token $(m,n)$ is obtained from a trimmed mean across group members:
\begin{equation}
r^l_{m,n}=\mathrm{TrimMean}_{\gamma}\left(\{z^l_{j,n',k^*}\}_{j\neq m}\right),
\end{equation}
where $k^*$ and $n'$ are the best slot and spatial token for the current candidate under reciprocal matching. The trimming fraction $\gamma$ removes a small number of extreme group members before aggregation.

We additionally measure rank dispersion
\begin{equation}
d^l_{m,n}=\mathrm{MAD}\left(\{\rho^l_{j,n',k^*}\}_{j\neq m}\right),
\end{equation}
where MAD is median absolute deviation. The final rank gate is
\begin{equation}
g^l_{m,n}=\sigma(\alpha r^l_{m,n}-\beta d^l_{m,n}).
\label{eq:gate}
\end{equation}
This gate rewards candidates whose group support is both strong and consistently ranked. A visually salient region that appears only in one image tends to exhibit weak support or unstable ranks and is suppressed.

\subsection{Group-Aware Decoder}
The gated group slots and image features are fused using cross-attention:
\begin{equation}
H_m^l=\mathrm{Attn}(F_m^l,G^l,G^l),\quad
G^l=\{g^l_k\bar{s}^l_k\}_{k=1}^{K}.
\end{equation}
We concatenate $H_m^l$ with the local feature map and pass it through a top-down feature pyramid. The decoder uses skip connections from the first two PVT-v2 levels to preserve thin structures and object boundaries. Such explicit separation between global context and local evidence is also compatible with structured layout representations, heterogeneous anchors, and low-overhead visual adaptation \cite{lu2025bbox,feng2025dolphin,wang2025visionlora,feng2026dolphinv2}.

\subsection{Set-Permutation Training}
During training, two independent permutations of the same image group are processed. The segmentation loss is applied to both outputs and a permutation consistency loss is added:
\begin{equation}
\mathcal{L}_{perm}=\frac{1}{M}\sum_m\|P_m-\widetilde P_m\|_1.
\end{equation}
This objective does not change the ground truth; it simply penalizes implementation details that leak the arbitrary ordering of images into the prediction.

\subsection{Hard-Distractor Augmentation}
To increase group selectivity, training groups are augmented by inserting visually salient but non-recurring regions into a subset of images. The inserted region is never labeled as co-salient. A distractor suppression term penalizes predicted activation inside the inserted region:
\begin{equation}
\mathcal{L}_{dis}=\frac{1}{|D|}\sum_{x\in D}P(x),
\end{equation}
where $D$ denotes the pasted distractor area. This augmentation directly targets false agreement without requiring category labels. Evidence-focused benchmarks and adaptive-routing studies show that heterogeneous or distracting context can expose failure modes that average-case evaluation misses \cite{wang2025wilddoc,fu2025ocrbenchv2,lu2025certainty}.

\subsection{Objective}
The overall objective is
\begin{equation}
\mathcal{L}=\mathcal{L}_{seg}+\lambda_p\mathcal{L}_{perm}+\lambda_d\mathcal{L}_{dis}+\lambda_e\mathcal{L}_{edge}.
\end{equation}
Here $\mathcal{L}_{seg}$ is the standard BCE plus IoU-style dense loss, $\mathcal{L}_{edge}$ supervises the boundary head, and the remaining terms enforce set properties.

\section{Evaluation Protocol}
\subsection{Datasets}
We use CoCA, CoSal2015, and CoSOD3k, following their public train/test organization \cite{fan2020deeper,fan2022rethinking}. The model is trained only on the training split. No category names are supplied to the network.

\subsection{Metrics}
We report S-measure, max $F_\beta$, max $E_\phi$, and MAE using standard foreground-map definitions \cite{fan2017smeasure,fan2018emeasure}. In addition, we report:
\begin{itemize}[leftmargin=*,itemsep=1pt,topsep=1pt]
\item \textbf{Rank Stability (RS):} inverse normalized dispersion of the candidate rankings across group members.
\item \textbf{Distractor Suppression (DS):} reduction in false-positive area under hard-distractor insertion.
\item \textbf{Permutation Gap (PG):} output difference between multiple group orderings.
\item \textbf{Group Robustness (GR):} area under the performance curve as group size decreases.
\end{itemize}

\begin{figure*}[t]
\centering
\includegraphics[width=0.96\textwidth]{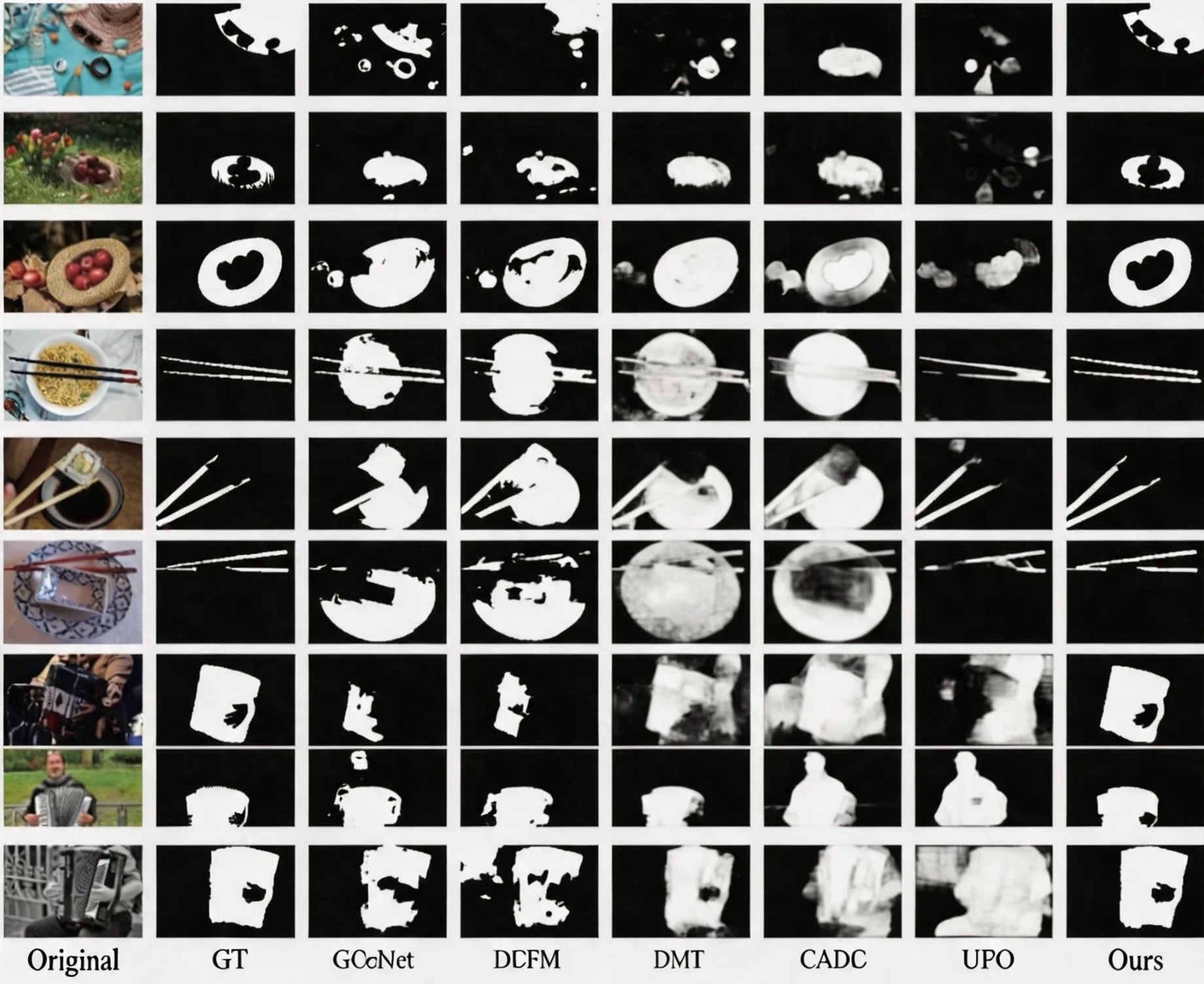}
\caption{Qualitative examples illustrating the visual diversity of co-salient objects across different image groups. The examples emphasize common-object appearance changes and background variation.}
\label{fig:qualitative}
\end{figure*}

\subsection{Baselines}
We compare with representative supervised Co-SOD methods: GCAGC, GCoNet, DCFM, DMT, GCoNet+, CONDA, LDRNet, and VCP \cite{zhang2020gcagc,fan2021gconet,yu2022democracy,li2023dmt,zheng2023gconetplus,li2024conda,piao2025ldrnet,wang2025vcp}. Published results from prior methods provide the comparison protocol and reference point for reproducible evaluation.

\subsection{Ablation Matrix}
We evaluate: (1) mean aggregation instead of trimmed rank aggregation; (2) no rank-dispersion gate; (3) no set slots and direct feature pooling; (4) no permutation loss; (5) no distractor augmentation; (6) single-scale reasoning; and (7) full RCSR. The aim is to identify which component changes group selectivity rather than simply increasing parameter count. This decomposition mirrors the diagnostic spirit of recent multimodal benchmarks and adaptive systems, where localization, reasoning, and robustness are evaluated separately \cite{tang2025mtvqa,shan2024mctbench,fu2025ocrbenchv2,wang2025wilddoc}.

\subsection{Group-Size and Order Tests}
At inference time, we evaluate each test group at 25\%, 50\%, 75\%, and 100\% of its available images. For each group, we also run five random permutations. Since the proposed representation is symmetric, the permutation gap should be near zero. Performance as group size changes reveals whether the model requires a large context window to succeed. Partial-context evaluation is especially relevant for models expected to infer structure from incomplete evidence and selective context \cite{wang2025pargo,shan2024mctbench,wang2025wilddoc,tang2025mtvqa,lu2025certainty,liu2026sleuth,yu2026ancientdoc}.

\subsection{Distractor Test}
For a fixed target image, we paste one or more salient objects sampled from unrelated groups into the background. We measure both the change in standard metrics and distractor suppression. A robust model should maintain the common target while limiting activation on inserted objects. Related work on object removal, distracting-token pruning, proactive focused perception, diffusion-quality probing, and structural error detection likewise emphasizes explicitly controlling irrelevant evidence \cite{sun2025attentive,li2026dtp,xue2026profocus,huang2026diffusionprobe,zhu2026textpecker}.

\begin{figure}[t]
\centering
\includegraphics[width=0.98\linewidth]{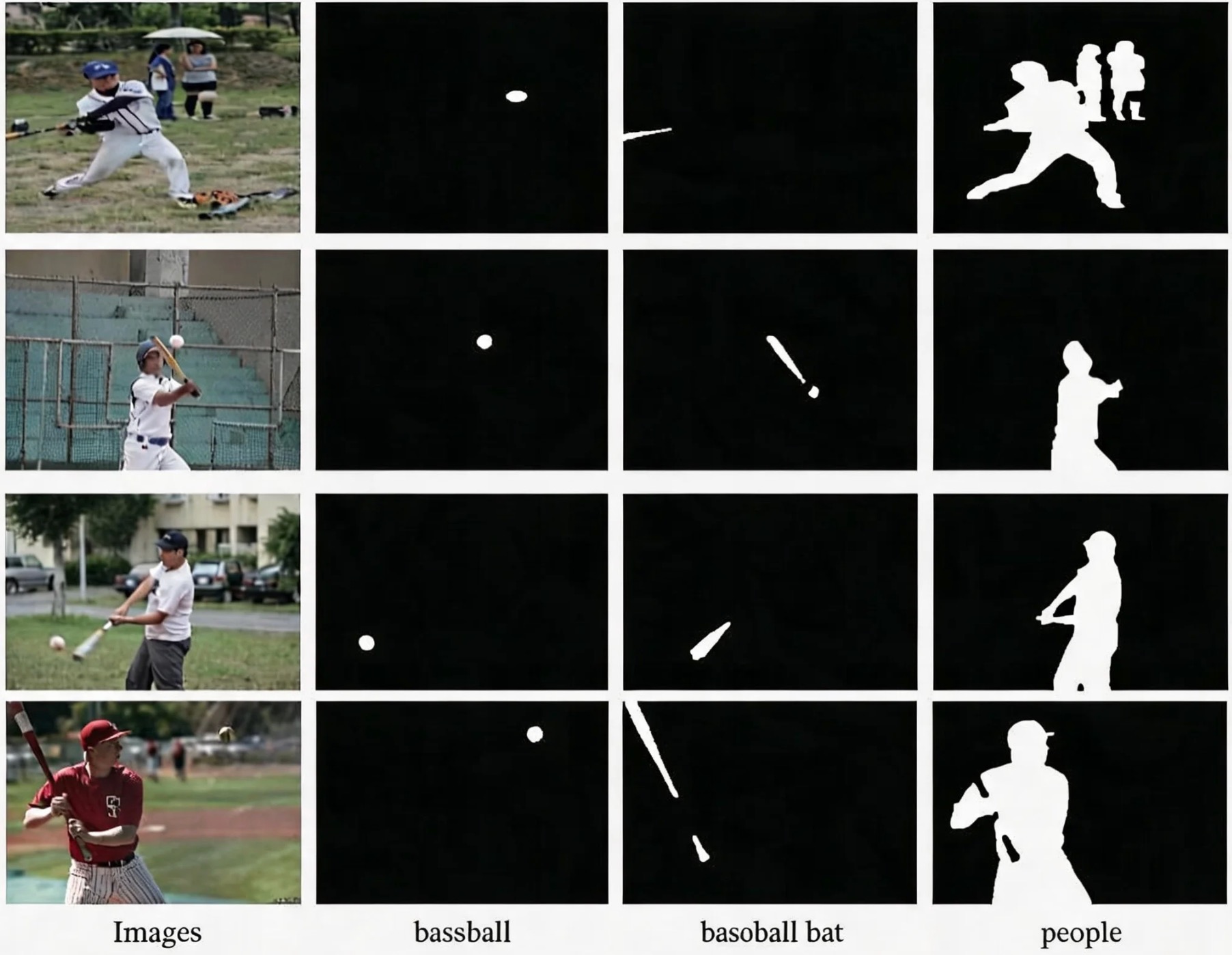}
\caption{Illustration of evidence selectability across an image group. Informative regions exhibit stronger recurrence and structural agreement, whereas distracting salient regions are less consistent across group members.}
\label{fig:selectability}
\end{figure}

\subsection{Cross-Dataset Transfer}
We train on one benchmark and evaluate directly on another without fine-tuning. This test is particularly important because the architecture does not depend on a dataset-specific vocabulary or external category list. We report both absolute performance and relative cross-domain drop. Cross-domain evaluation is motivated by the broader document and multimodal literature, where robustness to layout, language, and visual-domain variation is often a decisive test of learned representations \cite{feng2023unidoc,tang2024textsquare,yu2026ancientdoc}.

\section{Discussion and Limitations}
The rank-consistency view changes the role of group information. Instead of constructing one semantic label for the complete group, the model evaluates whether the ordering of candidate regions is stable across images. This makes the representation less dependent on absolute feature magnitudes and provides a direct diagnostic quantity for group agreement.

The main limitation is that rank-based aggregation can fail when every group member exhibits substantially different views of the target or when the target is extremely small. The method also requires supervised Co-SOD training and therefore does not address zero-shot inference. The broader literature on multimodal evidence selection and adaptive routing suggests several directions for reducing this supervision requirement \cite{wang2025pargo,wang2025visionlora,lu2025certainty,li2026dtp,jia2026meml}. Hard-distractor augmentation can improve selectivity but may still introduce synthetic artifacts that are not representative of natural backgrounds. These limitations should be quantified by the proposed stress tests rather than hidden in a single average benchmark score.

Recent CoSOD work emphasizes both generalization and strong training-free systems. CoMCS studies unseen-domain robustness with explicit style perturbation and semantic contrast, while TF-SSD explores a different training-free foundation-model pipeline \cite{guo2026comcs,he2026tfssd}. RCSR occupies a separate point in the design space: supervised, permutation-invariant, and centered on robust rank statistics instead of semantic prompting, proposal filtering, or counterfactual dataset synthesis.

\section{Reproducibility}
The implementation records the PVT-v2 checkpoint, input resolution, number of group slots $K$, projection dimension $d$, trimming fraction $\gamma$, rank-gate weights $(\alpha,\beta)$, feature-pyramid depth, optimizer, learning rate, training epochs, augmentation policy, and random seed.

We instantiate the network with a PVT-v2-B2 backbone, $K=8$ group slots, $\gamma=0.2$, and $(\alpha,\beta)=(2.0,1.0)$; the training schedule uses 80 epochs. Group-size evaluation uses four context fractions and repeated random permutations, while distractor evaluation inserts salient objects from unrelated groups under controlled appearance perturbations.

\section{Conclusion}
We presented Rank-Consistent Set Reasoning, a supervised Co-SOD framework that represents an image group as an unordered set and uses robust ranking statistics to separate recurrent foreground evidence from unstable salient regions. The method avoids category prediction, language prompting, open-vocabulary detection, and external segmentation. A permutation-invariant set encoder, rank-dispersion gate, and group-aware dense decoder form the central model. The proposed evaluation extends standard Co-SOD benchmarks with tests of order invariance, group-size robustness, distractor rejection, and cross-dataset transfer. The evaluation is intended to quantify the standard detection quality and the proposed set-level robustness properties.

{\small
\bibliographystyle{ieeenat_fullname}
\bibliography{references}
}

\end{document}